\documentclass[a4paper,11pt]{article}
\usepackage[top=4cm, bottom=4cm, left=3cm, right=3cm]{geometry}
\usepackage{amssymb}
\usepackage{amsmath}
\usepackage{amsthm}
\usepackage{mathrsfs} 
\usepackage[usenames,dvipsnames]{pstricks}
\usepackage{euler}
\usepackage{euscript}
\usepackage{graphicx}
\usepackage{charter}
\usepackage{mdframed}
\usepackage{enumerate, subfigure, color}
\usepackage{url}

\usepackage[round]{natbib}

\usepackage{booktabs}

\usepackage{multirow}

\usepackage{array}
\newcolumntype{M}[1]{>{\centering\arraybackslash}m{#1}}
\newcolumntype{N}{@{}m{0pt}@{}}

\newcommand{\specialcell}[2][c]{%
  \begin{tabular}[#1]{@{}c@{}}#2\end{tabular}}

\newcommand{\Nystrom}[1]{{Nystr\"om}}

\providecommand{\nor}[1]{\lVert{#1}\rVert}
\providecommand{\ran}[1]{\operatorname{range}{(#1)}}
\providecommand{\tr}{\operatorname{Tr}}

\newcommand{\R}{\mathbb R}

\newcommand{\hh}{\mathcal H}

\newcommand{\la}{\lambda}

\newcommand{\lspanc}[2]{\overline{\operatorname{span}\{#1~|~#2\}}}

\newcommand{\argmin}[1]{\mathop{\operatorname{argmin}}_{#1}}

\newcommand{\X}{{\cal X}}
\newcommand{\Y}{{\cal Y}}

\newcommand{\SNR}{\textrm{SNR}}

\newcommand{\eqals}[1]{{\begin{align*}#1\end{align*}}}
\newcommand{\eqal}[1]{{\begin{align}#1\end{align}}}
\newcommand{\bpr}{\begin{proof}}
\newcommand{\epr}{\end{proof}}
\newcommand{\be}{\begin{equation}}
\newcommand{\ee}{\end{equation}}

\newtheorem{definition}{Definition}
\newcommand{\bd}{\begin{definition}}
\newcommand{\ed}{\end{definition}}

\newcommand{\bi}{\begin{itemize}}
\newcommand{\ei}{\end{itemize}}

\newtheorem{ass}{Assumption}
\newcommand{\ba}{\begin{ass}}
\newcommand{\ea}{\end{ass}}

\newtheorem{remark}{Remark}
\newcommand{\br}{\begin{remark}}
\newcommand{\er}{\end{remark}}

\newtheorem{proposition}{Proposition}
\newcommand{\bp}{\begin{proposition}}
\newcommand{\ep}{\end{proposition}}

\newtheorem{lemma}{Lemma}
\newcommand{\blm}{\begin{lemma}}
\newcommand{\elm}{\end{lemma}}

\newtheorem{theorem}{Theorem}
\newcommand{\bt}{\begin{theorem}}
\newcommand{\et}{\end{theorem}}

\newtheorem{corollary}{Corollary}
\newcommand{\bcor}{\begin{corollary}}
\newcommand{\ecor}{\end{corollary}}

\makeatletter
\def\blfootnote{\xdef\@thefnmark{}\@footnotetext}
\makeatother

\title{\sffamily\LARGE NYTRO: When Subsampling Meets Early Stopping}
\author{Tomas Angles${}^{*1}$, \; Raffaello Camoriano${}^{\circ * 2,3}$, \; Alessandro Rudi${}^3$, \; Lorenzo Rosasco ${}^{1,3}$\\[3mm]
{\small ${}^1$ Massachusetts Institute of Technology 
and Istituto Italiano di Tecnologia}\\
{\small Laboratory for Computational and Statistical Learning,
Cambridge, MA 02139, USA}\\
{\small {\em \{ale\_rudi, lrosasco\}@mit.edu}}
\\[3mm]
{\small ${}^2$ Istituto Italiano di Tecnologia}\\
{\small iCub facility, Via Morego 30, Genova, Italy}\\
{\small {\em raffaello.camoriano@iit.it}}
\\[3mm]
{\small ${}^3$ Universit\`a degli Studi di Genova}\\
{\small DIBRIS, Via Dodecaneso 35, Genova, Italy}
\\[3mm]
}

\begin{document}
\maketitle
%

%

\begin{abstract}
Early stopping is a well known approach to reduce the time complexity for performing training and model selection of large scale learning machines. On the other hand,  memory/space (rather than time) complexity is the main  constraint in many applications, and randomized subsampling techniques have been proposed to tackle this issue. In this paper we ask whether early stopping and subsampling ideas can be  combined in a fruitful way. 
We consider the question in a least squares regression setting and propose a form of randomized iterative regularization based on early stopping and subsampling. In this context, we analyze the statistical and computational properties of the proposed method.
  Theoretical results are complemented and validated by a thorough experimental analysis.
\end{abstract}

\section{Introduction}
{\blfootnote{${}^*$The authors have contributed equally.}}{\blfootnote{${}^\circ$Corresponding author.}}Availability of large scale datasets requires the development of ever more efficient machine learning procedures.
A key feature towards scalability is being able to  tailor computational requirements to the generalization properties/statistical accuracy  allowed by the data.  In other words,  the precision with  which computations need to be performed  should be  determined  not only by the the amount, but also by the  quality  of the available data.

Early stopping,  known as iterative regularization in inverse problem theory  \citep{engl1996regularization,zhang2005boosting, bauer,earlyStopping,CapYao06}, provides a simple and sound implementation of this intuition. An empirical objective function is optimized in an iterative way with no explicit constraint or penalization and regularization is achieved  by suitably stopping the iteration. Too many iterations might lead to overfitting, while stopping too early might result in oversmoothing \citep{zhang2005boosting, bauer,earlyStopping,CapYao06}. Then, the best stopping rule arises from a form of bias-variance trade-off \citep{hastie2001elements}. Towards the discussion in the paper, the key observation is that the number of iterations controls at the same time the computational complexity as well as the statistical properties of the obtained learning algorithm \citep{earlyStopping}. Training and model selection can hence be performed with often considerable gain in time complexity.

Despite these nice properties, early stopping procedures often share the same space complexity requirements,
hence bottle necks, of other methods, such as those based on variational  regularization {\em \`a la Tikhonov} \citep[see][]{TikhonovOriginal,RidgeRegression}. A natural way to tackle these issues is to consider randomized subsampling/sketching approaches. Roughly speaking, these methods 
achieve memory and time savings by reducing the size of the problem in a stochastic way \citep{conf/icml/SmolaS00,conf/nips/WilliamsS00}. Subsampling methods
are typically used successfully together with penalized regularization. In particular, they are popular 
in the context of kernel methods, where they are often referred to as \Nystrom{} approaches and provide
one of the main methods towards large scale extensions \citep{conf/icml/SmolaS00,conf/nips/WilliamsS00,Zhang:2008:INL:1390156.1390311,conf/nips/KumarMT09,conf/icml/LiKL10,conf/nips/DaiXHLRBS14,HuangASSR14,conf/icml/SiHD14}.

In this paper, we ask whether early stopping and subsampling methods can be fruitfully combined. With the context of kernel methods in mind,  we propose and study NYTRO (NYstr\"om iTerative RegularizatiOn),  a simple algorithm combining these two ideas. After recalling  the properties and advantages of different regularization approaches in Section~\ref{sect:setting}, in Section~\ref{sect:proposedAlgorithm} we present in detail NYTRO and our main result, the characterization of its generalization properties. 
In particular, we analyze the conditions under which it attains the same statistical properties of subsampling and early stopping. Indeed, our study shows that while both techniques share similar, optimal, statistical properties, they are computationally advantageous in different regimes and NYTRO outperforms early stopping in the appropriate regime, as discussed in Section~\ref{sect:discussion}.
The theoretical results are validated empirically in Section~\ref{sect:experiments}, where NYTRO is shown to provide competitive results even at a fraction of the computational time, on a variety of benchmark datasets.

\section{Learning and Regularization}\label{sect:setting}
In this section we introduce the problem of learning in the fixed design setting and discuss different regularized learning 
approaches, comparing their statistical and computational properties. This section is  a survey that might be interesting in its own right, 
and reviews several results providing the context for the study in the paper.

\subsection{The Learning Problem}\label{sect:fix-setting}
We introduce the  learning setting we consider in the paper.
Let $\X = \R^d$ be the input space and $\Y \subseteq \R$  the output space. Consider  a {\em fixed design} setting \citep{conf/colt/Bach13} where the input points $x_1, \dots, x_n \in \X$ are fixed, while the outputs $y_1,\dots,y_n \in \Y$ are  given  by
$$ y_i = f_*(x_i) + \epsilon_i, \quad \forall \; i \in \{1,\dots, n\}$$
where $f_* : \X \to \Y$ is a fixed function and $\epsilon_1,\dots,\epsilon_n$ are random variables. The latter can be seen  seen as noise and are  assumed to be independently and identically distributed according to a probability distribution  $\rho$ with  zero mean and variance $\sigma^2$. 
In this context, the goal is  to minimize the {\em expected risk}, that is 
\eqal{\label{eq:ideal-problem}
 &\min_{f \in \hh} {\cal E}(f), ~~
 {\cal E}(f) = {\mathbb E} \frac{1}{n}\sum_{i=1}^n \left(f(x_i) - y_i\right)^2, ~~ \forall f \in \hh,
}
where $\hh$ is a space of functions, called {\em hypothesis space}.
In a real applications, $\rho$ and $f_*$ are unknown and  accessible only by means of a single realization $(x_1, y_1), \dots, (x_n, y_n)$ called {\em training set} and an approximate solution needs to be found.
The quality of a solution $f$ is measured by the {\em excess risk}, defined as
$$
R(f) = {\cal E}(f) - \inf_{v \in \hh} {\cal E}(v), \quad \forall f \in \hh.
$$
We next discuss estimation schemes to find a solution and compare their computational and statistical properties.

\subsection{From (Kernel) Ordinary Least Square to Tikhonov Regularization}

 A classical approach to derive an empirical solution to Problem~\eqref{eq:ideal-problem} is the so called {\em empirical risk minimization}
 \eqal{\label{eq:ols-problem}
f_\textrm{ols} = \argmin{f \in \hh} \frac{1}{n} \sum_{i=1}^n \left(f(x_i) - y_i\right)^2. 
}
In this paper, we are interested in the case where  $\hh$ is the reproducing kernel Hilbert space 
$$\hh = \lspanc{k(x,\cdot)}{x \in \X},$$
 induced by a positive definite kernel $k:\X \times \X \to \R$ \citep[see][]{schlkopf2002learning}.
In this case Problem~\eqref{eq:rls-problem} corresponds to the  {\em Kernel Ordinary Least Squares} (KOLS) 
and  has the closed form solution
\eqal{\label{eq:ols}
f_{\textrm{ols}}(x) & = \sum_{i=1}^n \alpha_{\textrm{ols}, i} k(x,x_i),\quad \alpha_{\textrm{ols}} =  K^{\dag} y,
}
for all $x \in \X$, where $(K)^\dag$ denotes the pseudo-inverse of the  $ \in \R^{n\times n}$  empirical kernel matrix $K_{ij} = k(x_i, x_j)$ and $y = (y_1, \dots, y_n)$. The cost for computing the coefficients $\alpha_{\textrm{ols}}$ is $O(n^2)$ in memory and $O(n^3 + q(\X) n^2)$ in time, where $q(\X) n^2$ is the cost for computing $K$ and $n^3$ the cost for obtaining its pseudo-inverse. Here $q(\X)$ is the cost of evaluating the kernel function. In the following,  we are concerned with the dependence on $n$ and hence view $q(\X)$
as a constant. 


The statistical properties of KOLS, and related methods,   can be characterized by suitable notions of {\em dimension}
that we recall next. 
 The simplest is  the {\em full} dimension, that is
 $$ d^* = \textrm{rank}\, K$$
which measures the degrees of freedom of the kernel matrix. This latter quantity might not be stable when 
$K$ is ill-conditioned. A more robust notion is provided by the {\em effective dimension}
$$ d_\textrm{eff}(\la) = \tr(K (K + \la n I)^{-1}),\quad \la > 0.$$
Indeed, the above quantity can be shown to be related 
to  the eigenvalue decay of $K$ \citep{conf/colt/Bach13,alaoui2014fast} and can be considerably smaller than $d^*$, as discussed in the following.
Finally,  consider 
\eqal{\label{eq:tilde-d}
\tilde{d}(\la) = n \max_i (K(K+\la n I)^{-1})_{ii},\quad \la > 0.
}
It is easy to see that the following inequalities hold,
$$ d_\textrm{eff}(\la) \leq \tilde{d}(\la)\leq 1/\la,\quad d_\textrm{eff}(\la) \leq d^*\leq n,\qquad \forall \la > 0.$$ 
Aside from the above notion of dimensionality, the statistical accuracy of empirical least squares solutions
depends on a natural form of signal to noise ratio defined next.
Note that  the function that minimizes the excess risk in $\hh$ is given by 
\eqals{
f_\textrm{opt} &= \sum_{i=1}^n \alpha_{\textrm{opt},i} k(x, x_i),\quad \forall x \in \X\\
\alpha_\textrm{opt} & =  K^\dag \mu, \quad \textrm{with}\;\; \mu = {\mathbb E} y.
}
Then, the signal to noise ratio is defined as 
\eqal{\label{eq:snr}
\SNR = \frac{\nor{f_\textrm{opt}}^2_\hh}{\sigma^2}.
}
Provided with the above definitions, we can recall a first basic results characterizing the statistical accuracy of KOLS.
\bt\label{thm:ols}
Under the assumptions of Section~\ref{sect:fix-setting}, the following equation holds,
$$ {\mathbb E} R(f_\textrm{ols}) = \frac{\sigma^2 d^*}{n}.$$
\et
The above result shows that  the excess risk of KOLS can be bounded in terms of the full dimension, the noise level and the number of points.  However, in general empirical risk minimization {\em does not} provide the best results and regularization is needed. We next recall this fact,  considering first Tikhonov regularization, that is the 
%
  {\em Kernel Regularized Least Squares} (KRLS) algorithm given by, 
\eqal{\label{eq:rls-problem}
\bar{f}_\la = \argmin{f \in \hh} \frac{1}{n} \sum_{i=1}^n \left(f(x_i) - y_i\right)^2 + \la\nor{f}^2_\hh. 
}
The above algorithm is a penalized empirical risk minimization problem. 
The  {\em representer theorem}  \citep{schlkopf2002learning} shows that  Problem~\eqref{eq:rls-problem} 
can be restricted to 
\eqal{\label{eq:repr-rls-problem}
\hh_n &= \{\sum_{i=1}^n \alpha_i k(\cdot, x_i)~|~ \alpha_1,\dots,\alpha_n \in \R\}.
}
Indeed, a direct computation shows that  the  solution of Problem~\eqref{eq:rls-problem} is 
\eqal{\label{eq:std-krls}
\bar{f}_\la(x) & = \sum_{i=1}^n \bar{\alpha}_{\la i} k(x,x_i),\quad \bar{\alpha}_\lambda =  (K + \la n I)^{-1} y,
}
for all $x \in \X$. The intuition that regularization can be beneficial is made precise by the following result comparing
KOLS and KRLS.

\bt\label{thm:rls}
Let $\la^* = \frac{1}{nSNR}$. The following inequalities hold, 
$$ {\mathbb E} R(\bar{f}_{\la^*}) \leq \frac{\sigma^2 d_\textrm{eff}(\la^*)}{n} <  \frac{\sigma^2 d^*}{n} = {\mathbb E} R(f_\textrm{ols}).$$
\et
We add a few comments. First, as announced, the above result quantifies the benefits of regularization.  Indeed,  it shows that there exists a $\la^*$ for which the expected excess risk of KRLS is smaller than the one of KOLS. As discussed  in Table~1 of \cite{conf/colt/Bach13}, if $d^* = n$ and the kernel is sufficiently ``rich'', namely universal \citep{micchelli2006universal},  then  
 $d_\textrm{eff}$ can be less than a fractional power of $d^*$, so that  $d_\textrm{eff} \ll d^*$ and 
 $$ {\mathbb E} R(\bar{f}_{\la^*}) \;\ll\; {\mathbb E} R(f_\textrm{ols}).$$ 
 Second, note that the choice of the regularization parameter depends on a  form of signal to noise ratio, which is usually unknown. In practice, a regularization path\footnote{The set of  solutions  corresponding to regularization parameters  in a discrete set $\Lambda \subset \R$.}  is computed and then a model selected or found by aggregation \citep{hastie2001elements}. Assuming the selection/aggregation
step to have negligible computational cost, the complexity of performing training {\em and } model selection is then
$O(n^2)$ in memory and $O\left(n^3|\Lambda|\right)$ in time. 
These latter requirements can become prohibitive when $n$ is large and the question is whether 
the same statistical accuracy of KRLS can be achieved while reducing time/memory requirements.\
\subsection{Early Stopping and \Nystrom{} Methods}\label{sect:from-tikh-to-early}
In this section, we first recall how early stopping regularization allows to achieve the same statistical 
accuracy of KRLS  with potential saving in time complexity. 
Then, we recall how subsampling ideas can be used  in the framework of Tikhonov regularization to reduce 
 the space complexity with no loss of statistical accuracy.


\paragraph{Iterative Regularization by Early Stopping} The idea is to consider the gradient descent minimization 
of Problem~\ref{eq:ols} for a fixed number of steps $t$. The corresponding  algorithm is then
\eqal{ \label{eq:iterative-reg}
\breve{f}_t(x) & = \sum_{i=1}^n \breve{\alpha}_{t,i} k(x_i, x),\\
\breve{\alpha}_t &= \breve{\alpha}_{t-1} - \frac{\gamma}{n} (K \breve{\alpha}_{t-1} - y),
}
where $\gamma < 1/\nor{K}$ and $\breve{\alpha}_0 = 0$. 
Note that in the above algorithm regularization is not achieved by explicit penalization or imposing constraints,
and the only tuning parameter is the number of steps. Indeed, as shown next, the latter controls at the same time the computational complexity and statistical accuracy of the algorithm. 
The following theorem compares the expected excess error of early stopping with the one of  KRLS.

\bt\label{thm:land-wrt-krls}
When $\gamma < 1/\nor{K}$ and $t \geq 2$ the following holds
$$
{\mathbb E} R\left(\breve{f}_{\gamma,t}\right) \leq c_t \; {\mathbb E} R\left(\bar{f}_{\frac{1}{\gamma t}}\right).  
$$
with $c_t = 4\left(1+\frac{1}{t-1}\right)^2 \leq 20$.
\et
The above theorem follows as a corollary of our main result given in Theorem~\ref{thm:main} and recovers
results essentially given in \cite{RaskuttiEarlyStopping}.  Combining the above result  with Theorem~\ref{thm:rls}, and setting $t^*=\frac{1}{\gamma \la^*}$, we have that 
$${\mathbb E} R\left(\breve{f}_{\gamma,t^*}\right) \approx {\mathbb E} R\left(\bar{f}_{\la^*}\right) \leq {\mathbb E} R\left(f_\textrm{ols}\right).
$$
The statistical accuracy of early stopping is essentially the same as KRLS and can be vastly better than a na\"{i}ve ERM approach. Note that the cost of computing the best possible solution with early stopping 
is $O(n^2t^*) = O(n^3 \SNR)$. Thus, the computational time of early stopping is proportional to the signal to noise ratio. Hence, it could be much better than  KRLS for noisy problems, that is when $\SNR$ is small. The main bottle neck of early stopping regularization is that it has the same space requirements of KRLS. Subsampling approaches have been proposed to tackle this issue.
\paragraph{Subsampling and Regularization} 
Recall that  the solution of the standard KRLS problem belongs to $\hh_n$. A basic idea \citep[see][]{conf/icml/SmolaS00}
is to consider  {\em \Nystrom{} KRLS} (NKRLS), restricting   Problem~\eqref{eq:rls-problem} to  a subspace $\hh_m \subseteq \hh_n$ defined as
\eqal{\label{eq:Hm}
\hh_m = \{\sum_{i=1}^m c_i k(\cdot,\tilde{x}_i) | c_1,\dots,c_m \in \R \}.
}
Here $M = \{\tilde{x}_1, \dots, \tilde{x}_m\}$ is a subset of the training set and  $m\le n$.
It is easy to see that the corresponding solution is given by 
\eqal{\label{eq:nys-sol}
\tilde{f}_{m,\la}(x) &= \sum_{i=1}^m (\tilde\alpha_{m,\la})_i k(x,\tilde{x}_i), \\
\tilde{\alpha}_{m,\la} &= (K_{nm}^\top K_{nm} + \la n K_{mm})^\dag K_{nm}^\top y,
} 
for all $x \in \X$, where $(\cdot)^\dag$ is the pseudoinverse, $\la > 0$, $K_{nm} \in \R^{n\times m}$ with $(K_{nm})_{ij} = k(x_i, \tilde{x}_j)$ and $K_{mm} \in \R^{m\times m}$ with $(K_{mm})_{i,j} = k(\tilde{x}_i, \tilde{x}_j)$. 
A more efficient formulation can also be derived. Indeed, we 
rewrite Problem~\eqref{eq:rls-problem}, restricted to $\hh_m$, as 
\eqal{
\tilde{\alpha}_{m,\la} & = \argmin{\alpha \in \R^m} \nor{K_{nm}\alpha - y}^2 + \la \alpha^\top K_{mm} \alpha\\
\label{eq:new-nys-problem}& = R \argmin{\beta \in \R^k} \nor{K_{nm}R\beta - y}^2 + \la \nor{\beta}^2
}
where in the last step we performed the change of variable $\alpha = R\beta$ where $R \in \R^{m\times k}$ is a matrix such that $R R^\top = K_{mm}^\dag$ and $k$ is the rank of $K_{mm}$. Then, we can obtain the following closed form expression,
\eqal{\label{eq:char-nyst}
\tilde{\alpha}_{m,\la} = R(A^\top A + \la n I)^{-1} A^\top y.
}
(see Prop.~\ref{prop:eq-nyst} in Section~\ref{sect:us-results} of the appendix for a complete proof).
This last formulation is convenient because  it is possible to compute $R$ by $R = S T^{-1}$ where $K_{mm} = S D$ is the economic QR decomposition of $K_{mm}$, with $S \in \R^{m \times k}$ such that $S^\top S = I$, $D \in \R^{k\times m}$ an upper triangular matrix and $T \in \R^{k \times k}$ an invertible triangular matrix that is the Cholesky decomposition of $S^\top K_{mm} S$. Assuming $k \approx m$, the complexity of \Nystrom{} KRLS is then $O(nm)$ in space  and $O(nm^2 + m^3|\Lambda|)$ in time. The following known result establishes the statistical accuracy of the solution obtained by suitably choosing the points in $M$.
%
\bt[Theorem 1 of \cite{conf/colt/Bach13}]\label{thm:nys}
Let $m \leq n$ and $M = \{\tilde{x}_1,\dots, \tilde{x}_m\}$ be a subset of the training set uniformly chosen at random. Let $\tilde{f}_{m,\la}$ be as in Equation~\eqref{eq:nys-sol} and $\bar{f}_{\la}$ as in Equation~\eqref{eq:std-krls} for any $\la > 0$. Let $\delta \in (0,1)$, when
$$m \geq \left(\frac{32 \tilde{d}(\la)}{\delta} + 2\right)\log \frac{\nor{K} n}{\delta \la}$$
with $\tilde{d}(\la) = n \sup_{1\leq i \leq n} (K(K+\la n I)^{-1})_{ii}$, then the following holds
$${\mathbb E}_M {\mathbb E} R\left(\tilde{f}_{m,\la}\right) \leq (1+4\delta) {\mathbb E} R\left(\bar{f}_{\la}\right).  
$$
\et
The above result shows that the space/time complexity of   NKRLS can be  adaptive to the statistical properties of the data
while preserving the same statistical accuracy of KRLS. Indeed, using  Theorem~\ref{thm:rls}, we have that
$${\mathbb E}_M {\mathbb E} R\left(\tilde{f}_{m,\la^*}\right) \approx {\mathbb E} R\left(\bar{f}_{\la^*}\right) < {\mathbb E} R\left(f_\textrm{ols}\right),$$
requiring $O(n \tilde{d}(\la^*)\log\frac{n}{\la^*})$ in memory and $O(n \tilde{d}(\la^*)^2(\log\frac{n}{\la^*})^2)$ in time. Thus, NKRLS is more efficient with respect to KRLS when $\tilde{d}(\la^*)$ is smaller than $\frac{n}{\log\frac{n}{\la^*}}$, that is when the problem is mildly complex.

Given the above discussion it is natural to ask whether subsampling and early stopping ideas can be fruitfully 
combined. Providing a positive answer to this question is the main contribution of this paper that we discuss next.

\section{Proposed Algorithm and Main Results}\label{sect:proposedAlgorithm}
We begin by describing the proposed algorithm incorporating  the \Nystrom{} approach described above
in  iterative regularization by early stopping.  The intuition is that the algorithm thus obtained could have memory and time complexity adapted to the statistical accuracy allowed by the data, while automatically computing the whole regularization path. Indeed, this intuition is then confirmed through a statistical analysis of the corresponding excess risk. Our result indicates in which regimes
 KRLS, NKRLS, Early Stopping and NYTRO are  preferable.

\subsection{The Algorithm}

 NYTRO is obtained considering a finite number of iterations of the  gradient descent minimization of the empirical 
 risk in Problem~\eqref{eq:ols-problem} over the space in Equation~\eqref{eq:Hm}.
 The algorithm thus obtained is given by, 
\eqal{\label{eq:nytro-sol}
 \hat{f}_{m,\gamma,t}(x) &= \sum_{i=1}^m (\hat{\alpha}_{m,t})_i k(\tilde{x}_i, x), \\
 \hat{\beta}_{m,\gamma,t} &= \hat{\beta}_{m,t-1} - \frac{\gamma}{n} R^\top (K_{nm}^\top(K_{nm}\hat{\beta}_{m,t-1} - y)), \\
 \hat{\alpha}_{m,\gamma,t} &= R\beta_{m,t},
} 
for all $x \in \X$, where $\gamma = 1/(\sup_{1\leq i \leq n} k(x_i,x_i))$ and $\hat{\beta}_{m,0} = 0$.
Considering that the cost of computing $R$ is $O(m^3)$, the total cost for the above algorithm is $O(nm)$ in memory and $O(nmt + m^3)$ in time.

In the previous section, we have seen that NKRLS has an accuracy comparable to the one of the standard KRLS under a suitable choice of $m$. We next show that, under the same conditions, the accuracy of NYTRO is comparable with the ones of KRLS and NKRLS, for suitable choices of $t$ and $m$.

\begin{figure*}[t]
\centering
\includegraphics[width=1\linewidth]{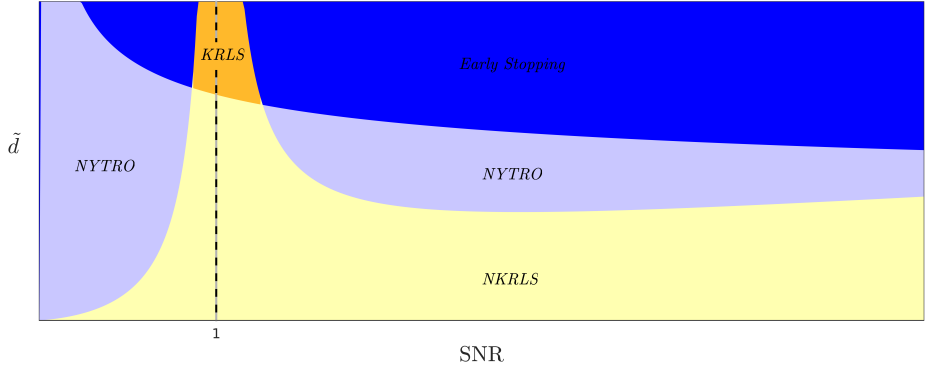}
\caption{The graph represents the family of learning problems parametrized by the dimensionality $\tilde{d}$ and the signal-to-noise ratio $\SNR$ (see Equations~\ref{eq:tilde-d},~\ref{eq:snr}). The four regions represent the regimes where different some algorithm are  faster than the others. Purple: NYTRO is faster, Blue: Early Stopping is faster, Orange: KRLS is faster, Yellow: NKRLS is faster -- see Section~\ref{sect:discussion}.}
\label{fig:areaPlotv2_WIDE}
\end{figure*}

\subsection{Error Analysis}
We next establish excess risk bounds for NYTRO by providing a direct comparison  
with  NKRLS and KRLS.
\bt[NYTRO and NKRLS] \label{thm:main}
Let $m \leq n$ and $M$ be a subset of the training set. Let $\hat{f}_{m,\gamma,t}$ be the NYTRO solution as in Equation~\eqref{eq:nytro-sol}, $\tilde{f}_{m,\frac{1}{\gamma t}}$ the NKRLS solution as in Equation~\eqref{eq:nys-sol}. When $t \geq 2$ and $\gamma < \nor{K_{nm} R}^2$ (for example $\gamma = 1/\max_i k(x_i,x_i)$) the following holds
$$
{\mathbb E} R\left(\hat{f}_{m,\gamma,t}\right) \leq c_t\; {\mathbb E} R\left(\tilde{f}_{m,\frac{1}{\gamma t}}\right).  
$$
with $c_t = 4\left(1+\frac{1}{t-1}\right)^2 \leq 16$.
\et
Note that the above result 
holds for any $m \leq n$ and any selection strategy of the \Nystrom{} subset $M$. 
The proof of Theorem~\ref{thm:main} is different from the one of Theorem~\ref{thm:nys} and is based 
only on geometric properties of the estimator and tools from spectral theory and inverse problems \citep[see][]{engl1996regularization}. 
In the next corollary we compare  NYTRO and KRLS, by combining Theorems~\ref{thm:nys} and ~\ref{thm:main}, hence considering $M$ to be chosen uniformly at random from the training set.
\bcor\label{cor:uniform}
Let $t \geq 2$, $\gamma = 1/\nor{K}$, $\delta \in (0,1)$ and $m$ be chosen as 
$$m \geq \left(32 \frac{\tilde{d}(1/(\gamma t))}{\delta} + 2\right)\log \frac{n\nor{K} \gamma t}{\delta}.$$
Let $\bar{f}_{\frac{1}{\gamma t}}$ be the KRLS solution as in Equation~\ref{eq:std-krls} and $\hat{f}_{m,\gamma,t}$ be the NYTRO solution. When the subset $M$ is chosen uniformly at random from the training set, the following holds
$$ {\mathbb E}_M {\mathbb E} R\left(\hat{f}_{m,\gamma,t}\right) \leq c_{t,\delta}\; {\mathbb E} R\left(\bar{f}_{\frac{1}{\gamma t}}\right)
$$
where $c_{t,\delta} = 4\left(1+\frac{1}{t-1}\right)^2(1 + 4 \delta) \leq 80$.
\ecor
The above result shows that NYTRO can achieve essentially the same results as KRLS.  In the next section we compare NYTRO to the other regularization algorithms introduced so far, by discussing how their computational complexity adapts to the statical accuracy in the data. In particular, by parametrizing the learning problems with respect to their dimension and their signal-to-noise ratio, we characterize the regions of the problem space where one algorithm is more efficient than the others.

\subsection{Discussion}\label{sect:discussion}
In Section~\ref{sect:setting} we have compared the expected excess risk of different regularization algorithms. More precisely, we have seen that there exists a suitable choice of $\la$ that is $\la^* = \frac{1}{n\SNR}$, where $\SNR$ is the signal-to-noise ratio associated to the learning problem,  such that the expected risk of KRLS is smaller than the one of KOLS, and indeed potentially much smaller.
For this reason, in the other result, statistical accuracy of the other methods was directly compared to that of 
 KRLS with $\la=\la^*$.

We exploit these results to analyze the complexity of the algorithms with respect to the statistical accuracy allowed by the data. If we choose $m \approx \tilde{d}(\la^*)\log (n/\la^*)$ and $t = \frac{1}{\gamma \la^*}$, then combining Theorem~\ref{thm:rls}   with Corollary~\ref{cor:uniform} and with Theorem~\ref{thm:nys}, respectively, we see that 
the expected excess risk of both NYTRO and NKRLS is in the same order of the one of KRLS. Both algorithms have a memory requirement of $O(nm)$ (compared to $O(n^2)$ for KRLS), but they differ in their time requirement. 
For NYTRO we have $O(n\frac{\tilde{d}(\la^*)}{\la^*}\log\frac{n}{\la^*})$, while for NKRLS it is $O(n\tilde{d}(\la^*)^2(\log \frac{n}{\la^*})^2)$. Now note that $\tilde{d}(\la^*)$ by definition is bounded by
$$ d_\textrm{eff}(\la) \leq \tilde{d}(\la) \leq \frac{1}{\la},\quad \forall \la > 0,$$
thus, by comparing the two computational times, we can identify two regimes
\eqals{
\begin{cases}
d_\textrm{eff}(\la^*) \leq\; \tilde{d}(\la^*)\; \leq \frac{1}{\la^* \log \frac{n}{\la^*}}  & \implies \textrm{NKRLS faster}\\
\frac{1}{\la^* \log\frac{n}{\la^*}} \leq \tilde{d}(\la^*) \leq \frac{1}{\la^*}  & \implies \textrm{NYTRO faster}
\end{cases}
} 
To illustrate  the regimes in which different algorithms can be preferable from a computational point of view while achieving the same error as KRLS with $\la^*$ (see Figure~\ref{fig:areaPlotv2_WIDE}), it is useful to 
parametrize the family of learning problems with respect to the signal-to-noise ratio defined in Equation~\eqref{eq:snr} and to the dimensionality of the problem $\tilde{d} := \tilde{d}(\la^*)$ defined in Equation~\eqref{eq:tilde-d}. We choose $\tilde{d}$ as a measure of dimensionality with respect to $d_\textrm{eff}$, because $\tilde{d}$ directly affects the computational properties of the analyzed algorithms.
In Figure~\ref{fig:areaPlotv2_WIDE},
the parameter space describing the learning problems is 
 partitioned in regions given by the curve 
that separates the subsampling methods from the standard methods and the curve that separates the iterative from Tikhonov methods.

\begin{table}
\caption{Specifications of the Datasets Used in Time-accuracy Comparison Experiments. $\sigma$ is the Bandwidth of the Gaussian Kernel.}
\begin{center}
\begin{tabular}{M{5cm}M{1.3cm}M{1.3cm}M{1.0cm}M{1.0cm}N}
\toprule
{\em Dataset} & $n$ &  $n_{test}$ & $d$ & $\sigma$ &\\
\midrule
{\em InsuranceCompany} & 5822 & 4000 & 85 &      3 &\\ 
{\em Adult} & 32562 & 16282 & 123 &    6.6 &\\ 
{\em Ijcnn} & 49990 & 91701 & 22 &      1 &\\ 
{\em YearPrediction} & 463715 & 51630 & 90 &      1 &\\ 
{\em CovertypeBinary} & 522910 & 58102 & 54 &      1 &\\ 
\bottomrule\hline
\end{tabular}

\end{center}
\label{tab:dataSpec}
\end{table}

As illustrated in Figure~\ref{fig:areaPlotv2_WIDE}, NYTRO is preferable when $\SNR \leq 1$, that is when the problem is quite noisy. 
When  $\SNR > 1$, then NYTRO is faster when the dimension of the problem is sufficiently large.
Note that,  in particular,  the area of the NYTRO region on $\SNR > 1$ increases with $n$, and  the  curve $c_1$ is quite flat when $n$ is very large. 
On the opposite extremes we have early stopping and NKRLS. Indeed, one is effective when the dimensionality is very large, while the second when it is very small. There is  a peak around $\SNR \approx 1$ for which it seems that the only useful algorithm is NKRLS when the dimensionality is sufficiently large.  The only region where KRLS is more effective is when $\SNR \approx 1$ and the dimensionality is close to $n$. 

In the next section, the theoretical results are validated by an experimental analysis on benchmark datasets.
We add one remark first.

\begin{remark}[Empirical  parameter choices and regularization path]
Note that an important aspect that is not covered by Figure~\ref{fig:areaPlotv2_WIDE} is that iterative algorithms have the further desirable  property of computing  the regularization path. In fact, for  KRLS and NKRLS computations are  slowed by a factor of $|\Lambda|$, where $\Lambda$ is the discrete set of cross-validated $\lambda$s. This last aspect is very relevant in practice, because the optimal regularization parameter values are not known and need to be found via model selection/aggregation. 
\end{remark}

\begin{table*}
\caption{Time-accuracy Comparison on Benchmark Datasets.}
\begin{center}
{\tiny\begin{tabular}{M{1.7cm}M{0.7cm}M{1.2cm}M{2cm}M{2cm}M{2cm}M{2cm}N}
\toprule
{\em Dataset} & & \specialcell{{\em KOLS}} & \specialcell{{\em KRLS}} & \specialcell{{\em Early}\\ {\em Stopping}} & \specialcell{{\em NKRLS}} & \specialcell{{\em NYTRO}} & \\
\midrule
\multirow{3}{*}{\specialcell{{\em InsuranceCompany}\\ $ n = 5822 $ \\ $m = 2000$  }}
 & $Time \, (s)$ & \textbf{1.04} & 97.48 $\pm$ 0.77 & 2.92 $\pm$ 0.04 & 20.32 $\pm$ 0.50 &   5.49 $\pm$ 0.12 & \\ 
 & $RMSE$ & 5.179 & \textbf{0.4651 $\pm$ 0.0001} & \textbf{0.4650 $\pm$ 0.0002} &  {\bf 0.4651 $\pm$ 0.0003} &   {\bf 0.4651 $\pm$ 0.0003} &\\ 
 & $Par.$ & NA & 3.27e-04 & 494 $\pm$ 1.7 & 5.14e-04 $\pm$ 1.42e-04 & 491 $\pm$ 3 &\\ 
 \hline 
\multirow{3}{*}{\specialcell{{\em Adult}\\ $ n = 32562 $ \\ $m = 1000$  }}
 & $Time \, (s)$ & 112 & 4360 $\pm$ 9.29 & 5.52 $\pm$ 0.23 & 5.95 $\pm$ 0.31 &   {\bf 0.85 $\pm$ 0.05} &\\ 
 & $RMSE$ & 1765 & \textbf{0.645 $\pm$ 0.001}  & 0.685 $\pm$ 0.002 & 0.6462 $\pm$ 0.003 &  0.6873 $\pm$ 0.003 & \\ 
 & $Par.$ & NA & 4.04e-05 $\pm$ 1.04e-05 & 39.2 $\pm$ 1.1 & 4.04e-05 $\pm$ 1.83e-05 & 44.9 $\pm$ 0.3 & \\ 
 \hline 
\multirow{3}{*}{\specialcell{{\em Ijcnn}\\ $ n = 49990 $ \\ $m = 5000$  }}
 & $Time \, (s)$ & 271 	& 825.01 $\pm$ 6.81 	& 154.82 $\pm$ 1.24 & 160.28 $\pm$ 1.54 &  {\bf 80.9 $\pm$ 0.4} &\\ 
 & $RMSE$ & 730.62 	& 0.615 $\pm$ 0.002 & {\bf 0.457 $\pm$ 0.001}& 0.469 $\pm$ 0.003 &  {\bf 0.457 $\pm$ 0.001} &\\ 
 & $Par.$ & NA 	& 1.07e-08 $\pm$ 1.47e-08 & 489 $\pm$ 7.2 & 1.07e-07 $\pm$ 1.15e-07 & 328.7 $\pm$ 2.6 &\\ 
 \hline 
\multirow{3}{*}{\specialcell{{\em YearPrediction}\\ $ n = 463715 $ \\ $m = 10000$  }}
 & $Time \, (s)$ &  &  & & 1188.47 $\pm$ 36.7 & {\bf 887 $\pm$ 6} &\\ 
 & $RMSE$ & NA & NA & NA & {\bf 0.1015 $\pm$ 0.0002} & 0.1149 $\pm$ 0.0002 &\\ 
 & $Par.$ & &  &  & 3.05e-07 $\pm$ 1.05e-07 & 481 $\pm$ 6.1 &\\ 
 \hline 
\multirow{3}{*}{\specialcell{{\em CovertypeBinary}\\ $ n = 522910 $ \\ $m = 10000$  }}
 & $Time \, (s)$   &  &  & & 1235.21 $\pm$ 42.1 &  {\bf 92.69 $\pm$ 2.35} &\\ 
 & $RMSE$ & NA & NA & NA & 1.204 $\pm$ 0.008 &  {\bf 0.918 $\pm$ 0.006} &\\ 
 & $Par.$ & & & & 9.33e-09 $\pm$ 1.12e-09 & 39.2 $\pm$ 2.3 &\\ 
 \bottomrule
 \hline 
\end{tabular}
}
\label{tab:testSetComparison}
\end{center}
\end{table*}

\section{Experiments}\label{sect:experiments}
In this section we present an empirical evaluation of the NYTRO algorithm, showing regimes in which it provides a significant model selection speedup with respect to NKRLS and the other exact kernelized learning algorithms mentioned above (KOLS, KRLS and Early Stopping). We  consider the Gaussian kernel and  the subsampling of the training set points for kernel matrix approximation is performed uniformly at random. All experiments have been carried out on a server with 12 $\times$ 2.10GHz Intel$^\circledR$ Xeon$^\circledR$ E5-2620 v2 CPUs and 132 GB of RAM.

We compare the algorithms on the benchmark datasets reported in Table \ref{tab:dataSpec}\footnote{All the datasets are available at \url{http://archive.ics.uci.edu/ml} or \url{https://www.csie.ntu.edu.tw/~cjlin/libsvmtools/datasets/}}. In the table we also report the bandwidth parameter $\sigma$ adopted for the Gaussian kernel computation. Following \citep[]{conf/icml/SiHD14}, we measure performance by the root mean squared error (RMSE).  
Note that for the \texttt{YearPredictionMSD} dataset outputs are scaled in $\left[ 0,1 \right] $. 

For all the  algorithms, model selection is performed via hold-out cross validation, where the validation set is composed of 20\% of the training points chosen uniformly at random at each trial. We select the regularization parameter $\lambda$ for NKRLS between $100$ guesses logarithmically spaced in $ \left[  10^{-15} ,  1 \right] $, by computing the validation error for each model and choosing the $\lambda^*$ associated with the lowest error. NYTRO's regularization parameter is the number of iterations $t$. We select the optimal $t^*$ by considering the evolution of the validation error. 
As an early stopping rule, we choose an iteration such that the validation error ceases to be decreasing up to a 
given threshold chosen to be the  $5\%$ of the relative RMSE.
After model selection, we evaluate the performance on the test set. We report the results in Table \ref{tab:testSetComparison} and discuss them further below.

\begin{figure}
\centering
\includegraphics[width = 0.45\textwidth]{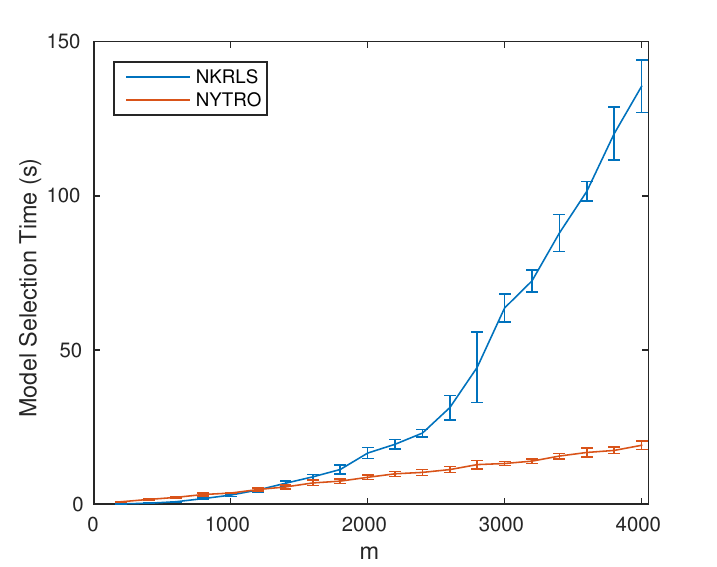}
\caption{Training Time of NKRLS and NYTRO on the \texttt{cpuSmall} Dataset as the Subsampling Level $m$ Varies Linearly Between 100 and 4000. Experiment With 5 Repetitions. Mean and Standard Deviation Reported.}
\label{fig:timeComparison}
\end{figure}

\paragraph{Time Complexity Comparison}
We start by showing how the time complexity changes with the subsampling level $m$, making NYTRO more convenient than NKRLS if $m$ is large enough. For example,  consider Figure \ref{fig:timeComparison}. We performed training on the \texttt{cpuSmall}\footnote{\url{http://www.cs.toronto.edu/~delve/data/datasets.html}} dataset ($n=6554,\, d=12$), with $m$ spanning between $100$ and $4000$ at $100$-points linear intervals. The experiment is repeated $5$ times, and we report the mean and standard deviation of the NYTRO and NKRLS model selection times. We consider $100$ guesses for $\lambda$, while the NYTRO iterations are fixed to a maximum of $500$. As revealed by the plot, the time complexity grows linearly with $m$ for NYTRO and quadratically for NKRLS. This is consistent with the time complexities outlined in Sections \ref{sect:setting} and \ref{sect:proposedAlgorithm} ($O(nm^2 + m^3)$ for NKRLS and $O(nmt + m^3)$ for NYTRO).

\paragraph{Time-accuracy Benchmarking}

We also compared the training time and accuracy performances for KRLS, KOLS, Early Stopping (ES), NKRLS and NYTRO, reporting the selected hyperparameter ($\lambda^*$ for KRLS and NKRLS, $t^*$ for ES and NYTRO), the model selection time and the test error in Table \ref{tab:testSetComparison}. All the experiments are repeated 5 times. The standard deviation of the results is negligible. Notably, NYTRO achieves comparable or superior predictive performances with respect to its counterparts in a fraction of the model selection time. In particular, the absolute time gains are most evident on large scale datasets such as \texttt{Covertype} and \texttt{YearPredictionMSD}, for which a reduction of an order of magnitude in cross-validation time corresponds to saving tens of minutes. Note that exact methods such as KOLS, KRLS and ES cannot be applied to such large scale datasets due to their prohibitive memory requirements. Remarkably, NYTRO's predictive performance is not significantly penalized in these regimes and can even be improved with respect to other methods, as in the \texttt{Covertype} case, where it requires 90\% less time for model selection. 

\section{Acknowledgements}
The work described in this paper is  supported by the Center for Brains, Minds and Machines (CBMM), funded by NSF STC award CCF-1231216, and FIRB project RBFR12M3AC, funded by the Italian Ministry of Education, University and Research.

\bibliographystyle{plainnat}
\bibliography{biblio.bib}

\appendix

\newpage


\section{Proofs}

\bpr[Proof of Theorem~\ref{thm:rls}]
By applying Prop.~\ref{prop:risk-dec} to the estimator of Equation~\ref{eq:ols} we have
$Q_\textrm{ols} = K^\dag K = P$. Now note that $P^2 = P$ by definition, $\tr(P) = d^*$ and that $P(I-P) = 0$, therefore
$${\mathbb E} R(f_\textrm{ols}) = \frac{\sigma^2}{n}\tr(P^2) + \frac{1}{n}\nor{P(I-P)\mu} = \frac{\sigma^2 d^*}{n}.$$
Now let $K = U \Sigma U^\top$ be the eigen-decomposition of $K$, with $U$ an orthonormal matrix and $\Sigma$ a diagonal matrix with $\sigma_1 \geq \dots \geq \sigma_n \geq 0$. Let $\bar{Q}_\la = (K + \la n I)^{-1} K$, $\beta = U^\top P\mu$ with $P = K^\dag K$ the projection operator on the range of $K$. By applying Prop.~\ref{prop:risk-dec} to the estimator of Equation~\eqref{eq:ols} and considering that $P(I -\bar{Q}_\la) = (I - \bar{Q}_\la)P$ and $I - \bar{Q}_\la = \la n (K + \la n I)^{-1}$, we have
\eqals{
{\mathbb E} R(\bar{f}_\la) & = \frac{\sigma^2}{n}\tr(\bar{Q}_\la^2) + \frac{1}{n}\nor{P(I - \bar{Q}_\la)\mu}^2 \\
& = \frac{\sigma^2}{n}\tr(\bar{Q}_\la^2) + \frac{1}{n}\nor{(I - \bar{Q}_\la)P\mu}^2 \\
& = \frac{1}{n}\sum_{i=1}^n \frac{\sigma^2 \sigma_i^2 + \la^2 n^2 \beta_i^2}{(\sigma_i + \la n)^2} \\
& = \frac{1}{n}\sum_{i=1}^{d^*} \frac{\sigma^2 \sigma_i^2 + \la^2 n^2 \beta_i^2}{(\sigma_i + \la n)^2}
} 
where the last step is due to the fact that $\sigma_i = \beta_i = 0$ for $i > d^*$.
By defining $\tau_i = \sigma_i^{-1/2} \beta_i$, and assuming $\la^2 n^2 \tau_i^2 \leq \sigma^2 \la n$, we have
\eqals{
	{\mathbb E} R(\bar{f}_{\la^*}) & =  \frac{1}{n}\sum_{i=1}^{d^*} \frac{\sigma^2 \sigma_i^2 + \la^2 n^2 \sigma_i \tau_i^2}{(\sigma_i + \la n)^2} \\
	& = \frac{1}{n}\sum_{i=1}^{d^*} \frac{\sigma_i}{\sigma_i + \la n} \frac{\sigma^2 \sigma_i + \la^2 n^2 \tau_i^2}{\sigma_i + \la n} \\
	& \leq \frac{\sigma^2}{n}\sum_{i=1}^{d^*} \frac{\sigma_i}{\sigma_i + \la n} = \frac{\sigma^2 d_{eff}(\la)}{n}.
}
Now note that $\la^2 n^2 \tau_i^2 \leq \sigma^2 \la n$ for any $\la \leq \frac{\sigma^2}{n \max_i \tau_i^2}$ and $\la^* \leq \frac{1}{n \max_i \tau_i^2}$, indeed 
$$\max_i \tau_i^2 \leq \sum_i \tau_i^2 = \nor{K^{-1/2} P \mu}_\hh^2 = \nor{f_{opt}}_\hh^2.$$
\epr

\bpr[Proof of Theorem~\ref{thm:land-wrt-krls}]
It is an application of Theorem~\ref{thm:main} when we select the whole training set ($m=n$) for the \Nystrom{} approximation. In that case, the expected excess risks of \Nystrom{} KRLS and NYTRO are just equal to the ones of KRLS and Early Stopping, indeed when $m=n$ we have that $K_{mm} = K_{nm} = K$. If we call $\bar{Q}_\la$ and $\tilde{Q}_{n,\la}$ the $Q$-matrices for the two algorithms (see Prop.~\ref{prop:risk-dec}) and $R$ such that $RR^\top = K_{mm}^\dag$, for any $\la > 0$ we have
\eqals{\bar{Q}_\la &= (K+\la n I)^{-1}K = (K K^\dag K + \la n I)^{-1}K K^\dag K  \\
& =  (K R R^\top K + \la n I)^{-1}K R R^\top K \\
& = K R ( R^\top K^2 R + \la n I)^{-1}R^\top K = \tilde{Q}_{n,\la}.
}
\epr

\bpr[Proof of Theorem~\ref{thm:main}]
In the following we assume without loss of generality that the selected points $\tilde{x}_1,\dots,\tilde{x}_m$ are the first $m$ points in the dataset.
In Prop.~\ref{prop:risk-dec} we have seen that the behavior of an algorithm in a fixed design setting is completely described by a matrix $Q = KC$ when the coefficients of the estimator are of the form $C y$. We now find the associated $Q$ for NYTRO, that is $\hat{Q}_{m,\gamma, t}$. By solving the recursion of Equation~\eqref{eq:nytro-sol}, we have for any $i \in \{1,\dots,n\}$
\eqals{
\hat{f}_{m,\gamma,t}(x_i) & = k_i^\top C y, \; \textrm{with } C = \begin{pmatrix}C_{m,\gamma,t}\\0_{(n-m) \times n}\end{pmatrix}, \\
C_{m,\gamma,t} & = \gamma \sum_{p=0}^{t-1} R(I - \gamma A^\top A)^p A^\top
}
with $A = K_{nm} R$ and $k_i = (k(x_i,x_1),\dots,k(x_i, x_n))$. Therefore, we have 
\eqals{
\hat{Q}_{m,\gamma,t}  &= K C = \gamma \sum_{p=0}^{t-1} K_{nm} R(I - \gamma A^\top A)^p A^\top \\
& = \gamma \sum_{p=0}^{t-1} A (I - \gamma A^\top A)^p A^\top.
}
{\bf Rewriting of $\hat{Q}_{m,\gamma,t}$.}
Now we rewrite $\hat{Q}_{m,\gamma,t}$ in a suitable form to bound the bias and variance errors.
First of all, we apply Prop.~\ref{prop:spectral} to $\hat{Q}_{m,\gamma,t}$. Let $f(\sigma) = \gamma\sum_{i=0}^{t-1} (1 - \gamma/n \sigma)^p$ with $\sigma \in [0 , n/\gamma]$, we have that
$$\hat{Q}_{m,\gamma,t} = Af(A^\top A)A^\top = f(A A^\top) A A^\top = g(A A^\top),$$
where $g(\sigma) = f(\sigma)\sigma$.
Now note that 
$$g(\sigma) = \gamma \sigma \sum_{i=0}^{t-1} (1 - \gamma/n \sigma)^p = 1 - (1 - \gamma/n \sigma)^t,$$
therefore we have
$$ \hat{Q}_{m,\gamma,t} = g(A A^\top) = I - (I - \gamma/n AA^\top )^t.$$

{\bf Bound of the Bias} Now we are going to bound the bias for NYTRO. Let $\la = 1/(\gamma t)$ and $Z = A A^\top$, then
\eqals{
B(\hat{Q}_{m,\gamma,t}) &= \frac{1}{n}\nor{P(I-\hat{Q}_{m,\gamma,t})\mu}^2 \\
& = \frac{1}{n}\nor{P(I-\frac{\gamma}{n} Z)^t\mu}^2 = \frac{1}{n}\nor{(I-\frac{\gamma}{n} Z)^tP\mu}^2\\
& = \frac{1}{n}\|(I-\frac{\gamma}{n} Z)^t (Z + \la n I)(Z + \la n I)^{-1} P \mu\|^2 \\
& \leq \frac{1}{n}q(A, \la n) \nor{(Z + \la n I)^{-1} P\mu}^2
}
and $q(A, \la n) =\nor{(I-\gamma/n A A^\top)^t (A A^\top + \la n I)}^2$. Note that the third step is due to the fact that $\ran Z \subseteq \ran{K} = \ran P$ and $Z$ is symmetric. Therefore, $Ph(Z) = h(Z)P$ as a consequence of Prop.~\ref{prop:spectral} for any spectral function $h$.
Let $\sigma_1,\dots, \sigma_n$ be the singular values of $Z$, we have
\eqals{
q\left(A, \frac{n}{\gamma t}\right) &= \sup_{i \in \{1,\dots,n\}} (1-\gamma/n\, \sigma_i)^{2t}\left(\sigma_i + \frac{n}{\gamma t}\right)^2 \\
& \leq \sup_{0 \leq \sigma \leq n/\gamma} (1-\gamma/n \,\sigma)^{2t}\left(\sigma + \frac{n}{\gamma t}\right)^2 \leq \frac{n^2}{\gamma^2 t^2}
}
Therefore we have
$$ 
B(\hat{Q}_{m,\gamma,t}) \leq \la^2 n \nor{(Z + \la n)^{-1} P \mu}^2.
$$

{\bf Bound of the Variance}
Let $t \geq 2$, $\la = \frac{1}{\gamma t}$, $r(\sigma) = (1-\gamma/n\, \sigma)^t $ and 
$$v(\sigma) = \sigma/(t-1) + \sigma(1 + r(\sigma)) - \la n (1 - r(\sigma)).$$
We have $v(\sigma) \geq 0$ for $0 \leq \sigma \leq n/\gamma$. Indeed, for $\la n < \sigma \leq n/\gamma$ we have $v(\sigma) \geq 0$, since $0 \leq r(\sigma) \leq 1$, while for $0 \leq \sigma \leq \la n$ we have
\eqals{
\la n (1 - r(\sigma)) & = \la n \left(1 - e^{-t \log \frac{1}{1 - \frac{\gamma \sigma}{n}}}\right) \leq \frac{n}{\gamma t} t \log \frac{1}{1 - \frac{\gamma \sigma}{n}} \\
& \leq \frac{n}{\gamma} \frac{\gamma/n\, \sigma}{1 - \gamma/n\, \sigma} \leq \frac{\sigma}{1 - \frac{1}{t}} = \frac{\sigma}{t - 1} +  \sigma \\
& \leq \frac{\sigma}{t-1} + \sigma(1 + r(\sigma)),
}
therefore $v(\sigma) \geq 0$.
Now let $0 \leq \sigma \leq n/\gamma$. Since $v(\sigma) \geq 0$, the function $w(\sigma) = v(\sigma)/(\sigma + \la n)$ is $w(\sigma) \geq 0$.
Now we rewrite $w$ a bit. First of all, note that
\eqals{
w(\sigma) = (2t-1)/(t-1) w_1(\sigma) - g(\sigma),
}
with $w_1(\sigma) = \sigma/(\sigma + \la n)$. 
The fact that $w(\sigma) \geq 0$ and that $g(\sigma) \geq 0$ implies that
$$ 
\left(\frac{2t-1}{t-1}\right)^2 w_1(\sigma)^2 \geq g(\sigma)^2. \quad \forall 0 \leq \sigma \leq \frac{n}{\gamma}, t \geq 2
$$
Let us now focus on $\tr (\hat{Q}_{\gamma t}^2)$. Let $Z = U\Sigma U^\top$ be its eigenvalue decomposition with $U$ an orthonormal matrix and $\Sigma = \textrm{diag}(\sigma_1,\dots,\sigma_n)$ with $\sigma_1 \geq \dots \geq \sigma_n \geq 0$,
\eqals{
\tr (\hat{Q}_{m,\gamma,t}^2) &= \tr(g^2(Z)) = \tr(U g^2(\Sigma) U^\top) = \tr(g^2(\Sigma))  \\
& = \sum_{i=1}^n g(\sigma_i)^2 \leq c_t\sum_{i=1}^n w_1(\sigma_i)^2 = c_t \tr (w_1(\Sigma)^2) \\
& = c_t \tr (U w_1(\Sigma)^2 U^\top) = c_t \tr (w_1(Z)^2) \\
& = c_t \tr (Z^2 (Z + \la n I)^{-2})
}
where we applied many times Prop.~\ref{prop:spectral} and the fact that the trace is invariant to unitary transforms.
Thus
$$ V(\hat{Q}_{m,\gamma,t},n) \leq \frac{\sigma^2}{n}\left(\frac{2t-1}{t-1}\right)^2\tr \left(Z \left(Z + n/(\gamma t) I\right)^{-1}\right)^2.
$$

{\bf The Expected Excess Risk for \Nystrom + KRLS}
The \Nystrom{} KRLS estimator with linear kernel is a function of the form 
\eqals{
\tilde{f}(x_i) &= k_i^\top C y, \quad \textrm{with } C = \begin{pmatrix}\tilde{C}_{m,\la}\\ 0_{(n-m)\times n} \end{pmatrix}, \\
\tilde{C}_{m,\la} & = R(A^\top A + \la n I)^\dag A^\top,
}
with $k_i = (k(x_i,x_1),\dots,k(x_i,x_n))$ for any $i \in \{1,\dots,n\}$. Now, by applying  Prop.~\ref{prop:spectral} we have
\eqals{
\tilde{Q}_{m,\la} & = KC = K_{nm}\tilde{C}_{m,\la} \\
& = A(A^\top A + \la n I)^{-1}A = AA^\top(A A^\top + \la I)^{-1}\\
& = Z(Z + \la n I)^{-1} 
}
Thus, we have
\eqals{
V(\tilde{Q}_{m,\la}) & = \frac{\sigma^2}{n}\tr (\tilde{Q}_{m,\la})^2 = \frac{\sigma^2}{n}\tr \left(Z \left(Z + \la n I\right)^{-1}\right)^2 \\
B(\tilde{Q}_{m,\la}) & = \frac{1}{n}\nor{P(I - Z \left(Z + \la n I\right)^{-1})\mu}^2 \\
& = \la^2 n  \nor{P(Z + \la n I)^{-1}\mu}^2 \\
& = \la^2 n \nor{(Z + \la n I)^{-1}P\mu}^2.
} 
where the last step is due to the same reasoning as in the bound for the bias of NYTRO.
Finally, by applying twice Prop.~\ref{prop:risk-dec} and calling $c_t = \left(\frac{2t-1}{t-1}\right)^2$, we have that
\eqals{
R(\hat{f}_{m,\gamma,t}) &= V(\hat{Q}_{m,\gamma,t},n) + B(\hat{Q}_{m,\gamma,t}) \\
&\leq c_t V(\tilde{Q}_{m,\frac{1}{\gamma t}},n) + B(\tilde{Q}_{m,\frac{1}{\gamma t}})  \\
&\leq c_t \left(V(\tilde{Q}_{m,\frac{1}{\gamma t}},n) + B(\tilde{Q}_{m,\frac{1}{\gamma t}})\right) \\
&= c_t R(\tilde{f}_{m,\frac{1}{\gamma t}})
}
for $\nor{Z} \leq n/\gamma$ and $t \geq 2$. Now the choice $\gamma = 1/(\max_{1\leq i \leq n} k(x_i,x_i))$ is valid, indeed
\eqals{
\gamma \nor{Z}^2 & = \gamma\nor{K_{nm} RR^\top K_{nm}^\top} = \gamma\nor{K_{nm} K_{mm}^\dag K_{nm}^\top}\\
& \leq \gamma\nor{K} \leq \gamma n \max_{1 \leq i \leq n} (K)_{ii} =  \gamma n \max_{1 \leq i \leq n} k(x_i, x_i),
}
where $\nor{K_{nm} K_{mm}^\dag K_{nm}^\top} \leq \nor{K}$ can be found in \cite{conf/colt/Bach13,alaoui2014fast}.
\epr

\bpr[Proof of Corollary~\ref{cor:uniform}]
Theorem~\ref{thm:main} combined with Theorem 1 of \cite{conf/colt/Bach13}.
\epr

\section{Some Useful Results}\label{sect:us-results}

\bp\label{prop:eq-nyst}
With the notation of Section~\ref{sect:from-tikh-to-early}, let $R \in \R^{m \times k}$ such that $K_{mm}^\dag = R R^\top$ and $A = K_{nm} R$. Then, for any $\la, m > 0$, $\tilde{\alpha}_{m,\la}$ is characterized by Equation~\ref{eq:char-nyst}.
\ep
\bpr
By Equation~7.7 of \cite{rifkin2003regularized} we have that
\eqals{
\tilde{\alpha}_{m,\la} &= K_{mm}^\dag K_{nm}^\top (K_{nm} K_{mm}^\dag K_{nm}^\top + \la n I)^{-1} y \\
&= R R^\top K_{nm}^\top (K_{nm} R R^\top K_{nm}^\top + \la n I)^{-1} y  \\
& = R A^\top (A A^\top + \la n I)^{-1} y\\
& = R(A^\top A + \la n I)^{-1} A^\top y,
}
where the last step is due to Prop.~\ref{prop:spectral}.
\epr
\bp\label{prop:risk-dec}
Let $k:\X\times \X \to \R$ be a kernel function on $\X$, $x_1,\dots,x_n$ be the given points and $y = (y_1,\dots,y_n)$ be the labels of the dataset.
For any function of the form $f(x) = \sum_{i=1}^n w_i k(x,x_i)$ with $w = C y$ for any $x \in \X$, with $C \in \R^{n \times n}$ independent from $y$, the following holds
$${\mathbb E}_y R(f) = \underbrace{\frac{\sigma^2}{n} \tr(Q^2)}_{\textrm{Variance } V(Q)} + \underbrace{\frac{1}{n}\nor{P(I - Q)\mu}^2}_{\textrm{Bias } B(Q)},$$
with $Q = K C \in \R^{n \times n}$, $K$ the kernel matrix, $\mu = {\mathbb E} y \in \R^n$ and $P = K^\dag K$ the projection operator on the range of $K$.
\ep
\bpr
A function $f \in \hh$ is of the form $f(x) = \sum_{i=1}^n \alpha_i k(x,x_i)$ for any $x \in \X$. If we compute it on a point of the dataset $x_i$, with $i \in \{1,\dots, n\}$, we have $f(x_i) = \sum_{j=1}^n \alpha_j k(x_i,x_j) = k_i^\top w$ with $w = Cy$ and $k_i = (k(x_i,x_1),\dots,k(x_i,x_n))$. Note that $K = (k_1,\dots, k_n)$.

{\bf Rewriting of {\cal E}, R for Fixed Design} We have 
\eqals{
{\cal E}(w) &= \frac{1}{n}\sum_{i=1}^n {\mathbb E} (k_i^\top w - y_i) = \frac{1}{n}\sum_{i=1}^n ({\mathbb E}\left(k_i^\top w - \mu_i\right)^2 \\
& \; - 2\left(k_i^\top w - \mu_i\right)\left(y_i - \mu_i\right) + (y_i - \mu_i)^2) \\
 &= \frac{1}{n}\sum_{i=1}^n (k_i^\top w - \mu_i)^2 + \frac{\sigma^2}{n} = \frac{\sigma^2}{n} + \frac{1}{n}\nor{K w - \mu}^2,
}
Now note that $PK = K$ and $(I-P)K = 0$, that $\nor{q}^2 = \nor{Pq}^2 + \nor{(I-P)q}^2$ for any $q \in \hh$ and that $\inf_{v \in \X} {\cal E}(v) = \sigma^2 + \nor{(I - P)\mu}^2$, then the excess risk can be rewritten as
\eqals{
R(w) & = \frac{1}{n}\nor{K w - \mu}^2 - \frac{1}{n}\nor{(I - P)\mu}^2\\
& = \frac{1}{n}\nor{P(K w - \mu)}^2 + \frac{1}{n}\nor{(I-P)(K w - \mu)}^2 \\
& \;\; - \frac{1}{n}\nor{(I - P)\mu}^2 = \frac{1}{n}\nor{P(K w - \mu)}^2.
}

{\bf Expected Excess Risk} We focus on the expectation of $R$ with respect to the dataset for linear functions that depend linearly on the observed labels $y$. Indeed we have
\eqals{
{\mathbb E} R(w) &= \frac{1}{n}{\mathbb E}\nor{P(KCy - P\mu)}^2 \\
& = \frac{1}{n}{\mathbb E}\nor{PQ (y-\mu) + P(I - Q)\mu}^2 \\
&= \frac{1}{n}{\mathbb E}\tr(Q (y-\mu)(y-\mu)^\top Q) + \frac{1}{n}\nor{P(I - Q)\mu}^2 \\
& \;\;\; - \frac{2}{n}{\mathbb E} (y-\mu)^\top QP(I - Q)\mu\\
&= \frac{1}{n}\tr(Q {\mathbb E}(y-\mu)(y-\mu)^\top Q) + \frac{1}{n}\nor{P(I - Q)\mu}^2\\
&= \frac{\sigma^2}{n} \tr(Q^2) + \frac{1}{n}\nor{P(I - Q)\mu}^2.
}
Here the third step is due to $\nor{a - b}^2 = \nor{a}^2 + \nor{b}^2 - 2 a^\top b$ and that $\nor{a}^2 = \tr (a a^\top)$, for any vector $a, b$. The last term in the third step vanishes due to the fact that $y-\mu$ is a zero mean random variable, moreover note that $({\mathbb E}(y-\mu)(y-\mu)^\top)_{ij} = {\mathbb E}(y_i - \mu_i)(y_j-\mu_j) = \sigma^2 \delta_{ij}$, therefore ${\mathbb E}(y-\mu)(y-\mu)^\top = \sigma^2 I$.
\epr

\bp[Spectral functions]\label{prop:spectral}
 Let $f, g: [0, T] \to \R$ be a continuous function and $A \in \R^{n \times n}$ symmetric with $\nor{A} \leq T$, for a $T > 0$, $n \geq 1$. Let $A = U \Sigma U^\top$ be its eigenvalue decomposition with $U \in \R^{n\times n}$ an orthonormal matrix, $U^\top U = U U^\top = I$ and $\Sigma$ a diagonal matrix, then
\eqals{
f(A) &= U f(\Sigma) U^\top,\\
 f(A) + g(A) &= (f + g)(A),\quad f(A)g(A) = (fg)(A)
 }
where $f(\Sigma) = \textrm{diag}(f(\sigma_1),\dots,f(\sigma_n))$.
Moreover, let $B \in \R^{n\times m}$ with $n,m \geq 1$, then
$$f(B^\top B) B^\top = B^\top f(B B^\top).$$
\ep

\end{document}